\documentclass[11pt,a4paper]{article}
\usepackage[hyperref]{acl2018}
\usepackage{times}
\usepackage{latexsym}
\usepackage{graphicx}
\usepackage{mdwlist}
\usepackage{amsmath}
\usepackage{color}
\usepackage{url}
\usepackage{booktabs}

\aclfinalcopy 
\def\aclpaperid{1303} 

\title{Automatic Estimation of Simultaneous Interpreter Performance}
\author{Craig Stewart,$^1$ Nikolai Vogler,$^1$ Junjie Hu,$^1$ Jordan Boyd-Graber,$^2$ Graham Neubig$^1$ \\ $^1$Language Technologies Institute, Carnegie Mellon University \\ $^2$CS, iSchool, UMIACS, LSC, University of Maryland}
\begin{document}
\maketitle

\begin{abstract}
Simultaneous interpretation, translation of the spoken word in real-time, is both highly challenging and physically demanding.
Methods to predict interpreter confidence and the adequacy of the interpreted message have a number of potential applications, such as in computer-assisted interpretation interfaces or pedagogical tools.
We propose the task of predicting simultaneous interpreter performance by building on existing methodology for quality estimation (QE) of machine translation output. 
In experiments over five settings in three language pairs, we extend a QE pipeline to estimate interpreter performance (as approximated by the METEOR evaluation metric) and propose novel features reflecting interpretation strategy and evaluation measures that further improve prediction accuracy.%
\footnote{https://github.com/craigastewart/qe\_sim\_interp} 
\end{abstract}

\section{Introduction}

Simultaneous Interpretation (SI) is an inherently difficult task that carries significant cognitive and attentional burdens. The role of the simultaneous interpreter is to accurately render the source speech in a given target language in a timely and precise manner. Interpreters employ a range of strategies, including generalization and summarization, to convey the source message as efficiently and reliably as possible~\cite{he16interpretese}. Unfortunately, the interpreter is pitched against the limits of human memory and stamina, and after only minutes of interpreting, the number of errors made by an interpreter begins to increase exponentially \cite{mosermercer98}.  

\begin{figure}[t]
\begin{center}
\includegraphics[width=0.4\textwidth]{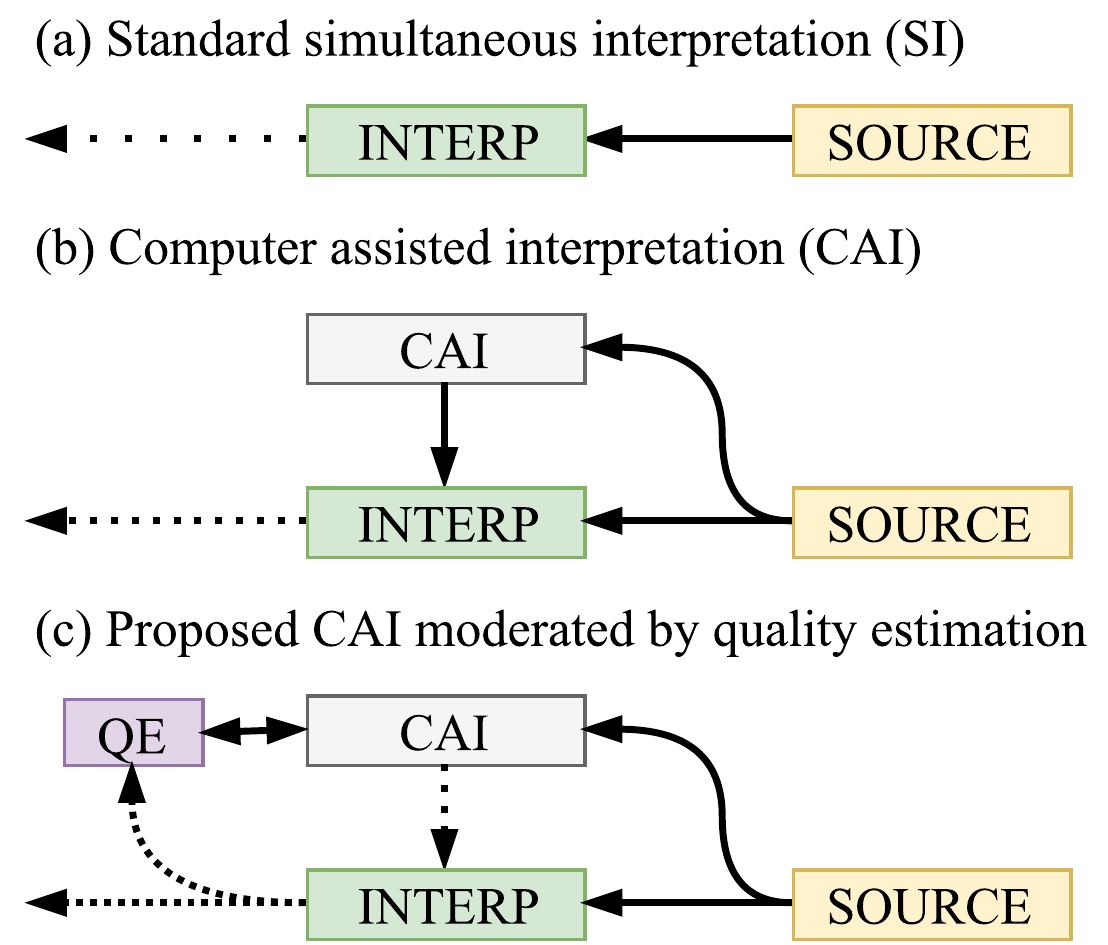}
\end{center}
\setlength{\abovecaptionskip}{-5pt}
\caption{Simultaneous interpretation scenarios} 

\label{fig:workflow}

\end{figure}

We examine the task of \emph{estimating simultaneous interpreter performance}: automatically predicting when interpreters are interpreting smoothly and when they are struggling. This has several immediate potential applications, one of which being in Computer-Assisted Interpretation (CAI). CAI is quickly gaining traction in the interpreting community, with software products such as InterpretBank~\cite{fantinouli16} deployed in interpreting booths to provide live and interactive terminology support. Figure~\ref{fig:workflow}(b) shows how this might work; both the interpreter and the CAI system receive the source message and the system displays assistive information in the form of terminology and informational support.

While this might improve the quality of interpreter output, there is a danger that these systems will provide \emph{too much} information and increase the cognitive load imposed upon the interpreter~\cite{fantinouli18}. Intuitively, the ideal level of support depends on current interpreter performance. The system can minimize distraction by providing assistance only when an interpreter is struggling. This level of support could be moderated appropriately if interpreter performance can be accurately predicted. Figure~\ref{fig:workflow}(c) demonstrates how our proposed quality estimation (QE) system receives and evaluates interpreter output, allowing the CAI system to appropriately lower the amount of information passed to the interpreter, maximizing the quality of interpreter output.


As a concrete method for estimating interpreter performance, we turn to existing work on QE for machine translation (MT) systems \cite{Specia10,specia-paetzold-scarton_ACL:2015}, which takes in the source sentence and MT-generated outputs and estimates a measure of quality.
In doing so, we arrive at two natural research questions:
\begin{enumerate*}
\item Do existing methods for performing QE on MT output also allow for accurate estimation of interpreter performance, despite the inherent differences between MT and SI?
\item What unique aspects of the problem of interpreter performance estimation, such as the availability of prosody and other linguistic cues, can be exploited to further improve the accuracy of our predictions?
\end{enumerate*}
The remainder of the paper describes methods and experiments on English-Japanese (EN-JA), English-French (EN-FR), and English-Italian (EN-IT) interpretation data attempting to answer these questions.

\section{Quality Estimation}
\label{sec:estimation}

\citet{Blatz:2004:CEM:1220355.1220401} first proposed the problem of measuring the quality of MT output as a prediction task, given that existing metrics such as BLEU~\cite{papineni02bleu} rely on the availability of reference translations to evaluate MT output quality, which aren't always available. As such, QE has since received widespread attention in the MT community and since 2012 has been included as a task in the Workshop on Statistical Machine Translation~\cite{callisonburch12wmt}, using approaches ranging from linear classifiers~\cite{ueffing2007word,luong14} to neural models~\cite{martins16wmt,martins17}.

QuEst++~\cite{specia-paetzold-scarton_ACL:2015} is a well-known QE pipeline that supports word-level, sentence-level, and document-level QE. Its effectiveness and flexibility make it an attractive candidate for our proposed task.
There are two main modules to QuEst++: a feature extractor and a learning module. The feature extractor produces an intermediate representation of the source and translation in a continuous feature vector. 
The goal of the learning module, given a source and translation pair, is to predict the quality of the translation, either as a label or as a continuous value.
This module is trained on example translations that have an assigned score (such as BLEU) and then predicts the score of a new example. QuEst++ offers a range of learning algorithms but defaults to Support Vector Regression for sentence-level QE.

\section{Quality Estimation for Interpretation}

The default, out-of-the-box, sentence-level feature set for QuEst++ includes seventeen features such as number of tokens in source/target utterances, average token length, $n$-gram frequency, etc. \cite{specia-paetzold-scarton_ACL:2015}.
While this feature set is effective for evaluation of MT output, SI output is inherently different---full of pauses, hesitations, paraphrases, re-orderings and repetitions. In the following sections, we describe our methods to adapt QE to handle these phenomena.


\subsection{Interpretation-specific Features}
\label{sec:interp-features}

To adapt QE to interpreter output, we augment the baseline feature set with four additional types of features that may indicate a struggling interpreter.

\paragraph{Ratio of pauses/hesitations/incomplete words:} \newcite{sridhar13simultaneous} propose that interpreters regularly use pauses to gain more time to think and as a cognitive strategy to manage memory constraints. An increased number of hesitations or incomplete words in interpreter output might indicate that an interpreter is struggling to produce accurate output. In our particular case, both corpora we use in experiments are annotated for pauses and partial renditions of words.


\paragraph{Ratio of non-specific words:} Interpreters often compress output by replacing or omitting common nouns to avoid specific terminology~\cite{sridhar13simultaneous}, either to prevent redundancy or to ease cognitive load. For example: ``The chairman explained the proposal to the delegates'' might be rendered in a target language as ``he explained it to them.'' To capture this, we include a feature that checks for words from a pre-determined seed list of pronouns and demonstrative adjectives.

\paragraph{Ratio of `quasi-'cognates:} In related language pairs, often words of a similar root are orthographically similar, for example ``artificial''(EN), ``artificiel''(FR) and ``artificiale''(IT). Likewise in Japanese, words adapted from English are transcribed in katakana script to indicate their foreign origin. Transliterated words in interpreted speech could represent facilitated translation by language proximity, or an attempt to produce an approximation of a word that the interpreter did not know. We include a feature that counts the number of words that share at least 50\% identical orthography (for EN, FR, IT) or are rendered in the interpreter transcript in katakana (JA).

\paragraph{Ratio of number of words:} We further include three features from QuEst++ that compare source and target length and the amount of transcribed punctuation. Information about utterance length makes sense in an interpreting scenario, given the aforementioned strategies of omission and compression of information. A list, for example, may be compressed to avoid redundancy or may be an erroneous omission \cite{barik94}.

\subsection{Evaluation Metric}


Novice interpreters are assessed for accuracy on the number of omissions, additions and the inaccurate renditions of lexical items and longer phrases \cite{altman94}, but recovery of content and correct terminology are highly valued. While no large corpus exists that has been manually annotated with these measures, they align with the phenomena that MT evaluation tries to solve.
One important design decision is which evaluation metric to target in our QE system.
There is an abundance of evaluation metrics available for MT including WER~\cite{Su92wer}, BLEU~\cite{papineni02bleu}, NIST~\cite{doddington02nist} and METEOR~\cite{denkowski:lavie:meteor-wmt:2014}, all of which compare the similarity between reference translations and translations.
Interpreter output is fundamentally different from any reference that we may use in evaluation because interpreters employ a range of economizing strategies such as segmentation, omission, generalization, and reformulation~\cite{riccardi05}. As such, measuring interpretation quality by some metrics employed in MT such as BLEU can result in artificially low scores~\cite{shimizu13iwslt}.
To mitigate this, we use METEOR, a more sophisticated MT evaluation metric that considers paraphrases and content-function word distinctions, and thus should be better equipped to deal with the disparity between MT and SI.
Better handling of these divergences for evaluation of interpreter output, or fine-grained evaluation based on measures from interpretation studies is an interesting direction for future work. 


\section{Data: Interpretation Corpora}

For our EN-JA language data we train the pipeline on combined data from seven TED Talks taken from the NAIST TED SI corpus \cite{shimizu13iwslt}. This corpus provides human transcribed SI output from three interpreters of low, intermediate and high levels of proficiency denoted B-rank, A-rank and S-rank respectively, with 559 utterances from each interpreter. The corpus also provides written translations of the source speech, which we use as reference translations when evaluating interpreter output using METEOR.

Our EN-FR and EN-IT data are drawn from the EPTIC corpus \cite{bernardini16},  which provides source and interpreter transcripts for speeches from the European Parliament (manually transcribed to include vocal expressions), as well as translations of transcripts of the source speech. The EN-FR and EN-IT datasets contain 739 and 731 utterances respectively.
While the EPTIC translations are accurate, they were created from an official transcript that differs significantly in register from the source speech. As a proxy for our experiments, we generated translations of the original speech using Google Translate, which resulted in much more qualitatively reliable METEOR scores than the EPTIC translations.

\section{Interpreter Quality Experiments}
\label{sec:experiments}


To evaluate the quality of our QE system, we use the Pearson's ~$r$ correlation between the predicted and true METEOR for each language pair~\cite{graham:2015:ACL-IJCNLP}.
As a baseline, we train QuEst++ on the out-of-the-box feature set (Section~\ref{sec:estimation}).

We use $k$-fold cross-validation individually on EN-JA, EN-FR, and EN-IT source-interpreter language pairs with a held-out development set and test set for each fold. For each experiment setting, we run the experiment for each fold (ten iterations for each set) and evaluate average Pearson's $r$ correlation on the development set.

In our baseline, we extract features based on the default QuEst++ sentence-level features (\textit{baseline}). We ablate baseline features via cross-validation and remove relating bigram, trigram, and punctuation frequency features in the source utterance, creating a more effective trimmed model (\textit{trimmed}).

Subsequently, we add our interpreter features (Section~\ref{sec:interp-features}) and arrive at our \textit{proposed} model. We then repeat each experiment using the test set data from each fold and compare the resulting average Pearson's $r$ scores.

\subsection{Results}



\begin{table}[t!]
\resizebox{\columnwidth}{!}{%
\begin{tabular}{l l l l}
  \toprule
  & baseline & trimmed & proposed \\
  \midrule
	EN-JA(B-rank) & 0.514 	 & 0.542   & \textbf{0.593} \\ 
	EN-JA(A-rank) & 0.487 	 & 0.554   & \textbf{0.591} \\
	EN-JA(S-rank) & 0.325 	 & 0.334   & \textbf{0.411} \\
	EN-FR 		  & 0.631 	 & 0.610   & \textbf{0.691} \\
	EN-IT 		  & 0.569 	 & 0.543   & \textbf{0.576} \\
  \bottomrule
\end{tabular}
}
\caption{Pearson's $r$ scores for predicted METEOR for baseline, trimmed and proposed feature sets on the test set (highest accuracy for each dataset indicated in bold).}
\label{tab:correlation}
\end{table}

Table~\ref{tab:correlation} shows our primary results comparing the baseline, trimmed, and proposed feature sets.
Our first observation is that, even with the baseline feature set, QE obtains respectable correlation scores, proving feasible as a method to predict interpreter performance.
Our trimmed feature set performs moderately better than the baseline for Japanese, and slightly under-performs for French and Italian. However, our proposed, interpreter-focused model out-performs in all language settings with notable gains in particular for EN-JA(A-Rank) (+0.104), achieving its highest accuracy on the EN-FR dataset.
Over all datasets, the gain of the proposed model is statistically significant at $p<0.05$ by the pairwise bootstrap~\cite{koehn04sigtest}.





\subsection{Analysis}

We further present two analyses: ablation on the full feature set and a qualitative comparison.  Table~\ref{tab:ablation} iteratively reduces the feature set by first removing the `quasi-'cognate feature (w/o cog); specific words (w/o spec); pauses, hesitations, and incomplete words (w/o fill); and finally sentence length and punctuation differences (w/o length).

\begin{table}[t!]
\resizebox{\columnwidth}{!}{%
\begin{tabular}{l l l l l}
  \toprule
 & w/o cog & w/o spec & w/o fill & w/o len \\
  \midrule
	EN-JA(B) 	  &+0.007& +0.012 	 & +0.016   & -0.053 \\ 
	EN-JA(A) 	  &-0.006& -0.011 	 & -0.012   & -0.031 \\
	EN-JA(S) 	  &-0.014& +0.001 	 & +0.004   & -0.061 \\
	EN-FR 		  &-0.013& -0.006 	 & +0.007   & -0.054 \\
	EN-IT 		  &-0.020& +0.002 	 & +0.020   & +0.005 \\
    \midrule
    Average 	  &-0.009& -0.001 	 & +0.007   & -0.039 \\
  \bottomrule
\end{tabular}
}
\caption{Relative difference in Pearson's $r$ scores for ablated features after removing cognates, specifics, fillers and length difference (cumulative ablation, left to right). Omission and addition are key features distinguishing SI from translation.}
\label{tab:ablation}
\end{table}

Relative difference in utterance length appears to aid Japanese and French above other languages. Cognates are particularly useful in EN-FR and EN-IT; this may be indicative of the corpus domain (European Parliament proceedings being rich in Latinate legalese) or of cognate frequency in those languages. In Japanese, cognates were more indicative of quality for the more skilled interpreter (S-rank).
While pauses and hesitations seem to aid the model in EN-FR and EN-IT, they appear to hinder EN-JA.

Below is a qualitative EN-IT example with METEOR score 0.079 (substantially lower than the average METEOR score across all datasets; 0.262). The baseline predicts a score of 0.127, while our model predicts 0.066:

\begin{small}
\vspace{3mm}
\noindent
{\bf SOURCE:} ``\textit{Will the Parliament grant President Dilma Rousseff, on the very first occasion after her groundbaking groundbreaking election and for no sound formal reason, the kind of debate that we usually reserve for people like Mugabe? So, I ask you to remove Brazil from the agenda of the urgencies.}'' 

\vspace{3mm}
\noindent
{\bf INTERP:} ``\textit{Ehm il Parlamento... dopo le elezioni... darem- darà spazio a un dibattito sul ehm sul caso per esempio del presidente Mugabe invece di mettere il Brasile all'ordine del giorno?}'' 

\vspace{3mm}
\noindent
{\bf GLOSS:} ``\textit{Ehm the Parliament... after the elections... we'll gi- will give way to a debate on the ehm on the case for example of President Mugabe instead of putting Brazil on the agenda?}''

\end{small}

\vspace{3mm}
Our model can better capture the issues in this example because it has many interpretation specific qualities (pauses, compression, and omission). This is an example in which a CAI system might offer assistance to an interpreter struggling to produce an accurate rendition.


\section{Conclusion}

We introduce a novel and effective application of QE to evaluate interpreter output, which could be immediately applied to allow CAI systems to selectively offer assistance to struggling interpreters.
This work uses METEOR to evaluate interpreter output, but creation of fine-grained measures to evaluate various aspects of interpreter performance is an interesting avenue for future work.

\section*{Acknowledgements}

This material is based upon work supported by the National Science Foundation under Grant No. 1748642 (CMU), 1748663 (UMD), and Graduate Research Fellowship No. DGE1745016.  We thank Leah Findlater and Hal Daum\'e III for useful feedback.

\bibliography{myabbrv,neubig}
\bibliographystyle{acl_natbib}

\end{document}